\documentclass[conference]{IEEEtran}
\IEEEoverridecommandlockouts

\usepackage{cite}
\usepackage{amsmath,amssymb,amsfonts}
\usepackage{algorithmic}
\usepackage{graphicx}
\usepackage{textcomp}
\usepackage{xcolor}

\usepackage{url}
\usepackage[normalem]{ulem}
\usepackage{booktabs}
\usepackage[table]{xcolor}
\usepackage{multirow}

\def\BibTeX{{\rm B\kern-.05em{\sc i\kern-.025em b}\kern-.08em
    T\kern-.1667em\lower.7ex\hbox{E}\kern-.125emX}}

\newcommand{\best}[1]{\colorbox{blue!18}{#1}}
\newcommand{\second}[1]{\colorbox{green!20}{#1}}
\begin{document}

\title{Simple Baselines for Fetal Brain Unsupervised Anomaly Detection%
\thanks{\(^{\ddagger}\) Co-first author. \hspace{1em} * Corresponding author: qiyuantian@tsinghua.edu.cn, arktisx@foxmail.com.}}

\author{
\IEEEauthorblockN{
Hongjia Yang\(^{1,\ddagger}\),
Mingxuan Liu\(^{1,\ddagger,*}\),
Yingqi Hao\(^{1,\ddagger}\),
Xiaotian Hu\(^{1}\),
Siyao Li\(^{1}\),
Xuguang Bai\(^{1}\),
Fenglin Jia\(^{2}\)\\
Yijin Li\(^{1}\),
Yifei Chen\(^{1}\),
Zihan Li\(^{1}\),
Juncheng Zhu\(^{2}\),
Jing Liu\(^{2}\),
Yi Liao\(^{2}\),
Ziyu Li\(^{3}\),
Haibo Qu\(^{2}\),
Qiyuan Tian\(^{1,*}\)
}
\IEEEauthorblockA{
\textit{\(^{1}\)School of Biomedical Engineering, Tsinghua Medicine, Tsinghua University, Beijing, China}\\
\textit{\(^{2}\)Department of Radiology, West China Second University Hospital, Sichuan University, Chengdu, China}\\
\textit{\(^{3}\) Nuffield Department of Clinical Neurosciences, University of Oxford, Oxford, UK}
}
}

\maketitle

\begin{abstract}
Unsupervised anomaly detection (UAD) offers a promising approach to fetal brain magnetic resonance imaging (MRI) analysis, where pathological data are scarce. Existing methods rely mainly on brain age prediction or image reconstruction, but are limited by noisy gestational age (GA) labels, lack of lesion localization, or poor cross-domain generalization. We propose FetalBUAD, a UAD framework based on voxel-level brain age estimation. A joint multi-task network predicts global brain age, voxel-level brain age maps, and brain segmentation to establish an internal biological brain age reference, reducing reliance on clinically recorded GA. Uncertainty-aware voxel-level brain age gap maps are then used for anomaly localization. Experiments on 1,317 internal fetal brain MRI volumes and an external benchmark demonstrate state-of-the-art anomaly detection performance and improved cross-domain generalization. FetalBUAD also generates anatomically interpretable maps that localize diverse fetal brain abnormalities. Our code is publicly available at: \url{https://github.com/birthlab/FetalBUAD}.
\end{abstract}

\begin{IEEEkeywords}
Fetal Brain MRI, Brain Age Prediction, Unsupervised Anomaly Detection.
\end{IEEEkeywords}

\section{Introduction}
Prenatal brain development is a critical phase that establishes the foundation for lifelong neurodevelopmental health \cite{natureHB}. Deviations from normal developmental trajectories are associated with long-term neurological disorders, including epilepsy \cite{epilepsy}, intellectual disabilities \cite{intellectual}, and autism spectrum disorder \cite{autism}. Accurate identification of atypical fetal brain development is essential for timely prenatal intervention and counseling. While ultrasound is the primary screening modality, fetal magnetic resonance imaging (MRI) provides superior soft-tissue contrast and a larger field of view \cite{fetalMRI,10.1007/978-3-032-05997-0_11}, enabling comprehensive assessment of complex intracranial structures.

Deep learning has shown potential for fetal brain abnormality diagnosis \cite{10.1016/j.jksuci.2023.101647}. However, developing robust supervised diagnostic models remains challenging. Pathological fetal MRI data are scarce due to the low prevalence of many congenital brain disorders, ethical constraints on data collection, and high expert annotation costs. Furthermore, fetal brain abnormalities vary substantially in anatomical presentation and severity, with many rare disorders represented by only a handful of cases. These challenges hinder supervised models from characterizing the full spectrum of fetal brain abnormalities and limit their clinical applicability.

Unsupervised anomaly detection (UAD) offers an alternative by learning normal fetal brain development without pathological annotations. One approach formulates anomaly detection as brain age prediction, using the discrepancy between predicted brain age and clinically recorded gestational age (GA) as an anomaly score \cite{single_slice,multi_slice,whole_brain,PANDA}. However, this strategy has two fundamental limitations. First, it cannot provide voxel-level localization because it estimates only a global age difference. Second, its reliability is constrained by inaccurate clinical GA labels. Although GA is commonly estimated using crown-rump length measurements from first-trimester ultrasound \cite{clinicalGA1}, measurement uncertainty and inter-subject variability in fetal growth can cause errors of up to $\pm$5--7 days \cite{clinicalGAerror}. Consequently, the brain age gap may reflect inaccurate clinical dating rather than genuine developmental abnormalities, reducing the reliability of age-gap-based anomaly detection.

Another major family of UAD methods detects abnormalities through discrepancies between input images and learned normal representations \cite{10540605,AE,DAE,sftp,iter} in image or feature space. These methods assume that models trained exclusively on normal data cannot faithfully reproduce abnormal structures, resulting in larger reconstruction or feature discrepancies in pathological regions. While effective for conspicuous intensity abnormalities, they may perform poorly when structural abnormalities cause subtle appearance changes. Moreover, learned normal representations depend on imaging characteristics and are susceptible to domain shifts across scanners, acquisition protocols, and institutions, limiting generalization.

To address these limitations, we propose \textbf{\texttt{FetalBUAD}}, an unsupervised fetal brain anomaly detection framework based on voxel-level biological brain age estimation. Rather than directly comparing predictions with potentially inaccurate clinical GA labels, \textbf{\texttt{FetalBUAD}} establishes an internal biological age reference through a joint multi-task network that predicts global brain age, voxel-level brain age maps, and brain segmentation \cite{UnetvolBAP,VBAP}. During inference, Monte Carlo (MC) Dropout \cite{MCdropout} estimates epistemic uncertainty to weight the voxel-level brain age gap map relative to the biological reference. Experiments on large-scale internal and external datasets demonstrate that \textbf{\texttt{FetalBUAD}} achieves state-of-the-art anomaly detection, superior cross-domain generalization, and anatomically interpretable lesion localization.

\begin{figure*}[!t]
	\centering
	\includegraphics[width=0.96\linewidth]{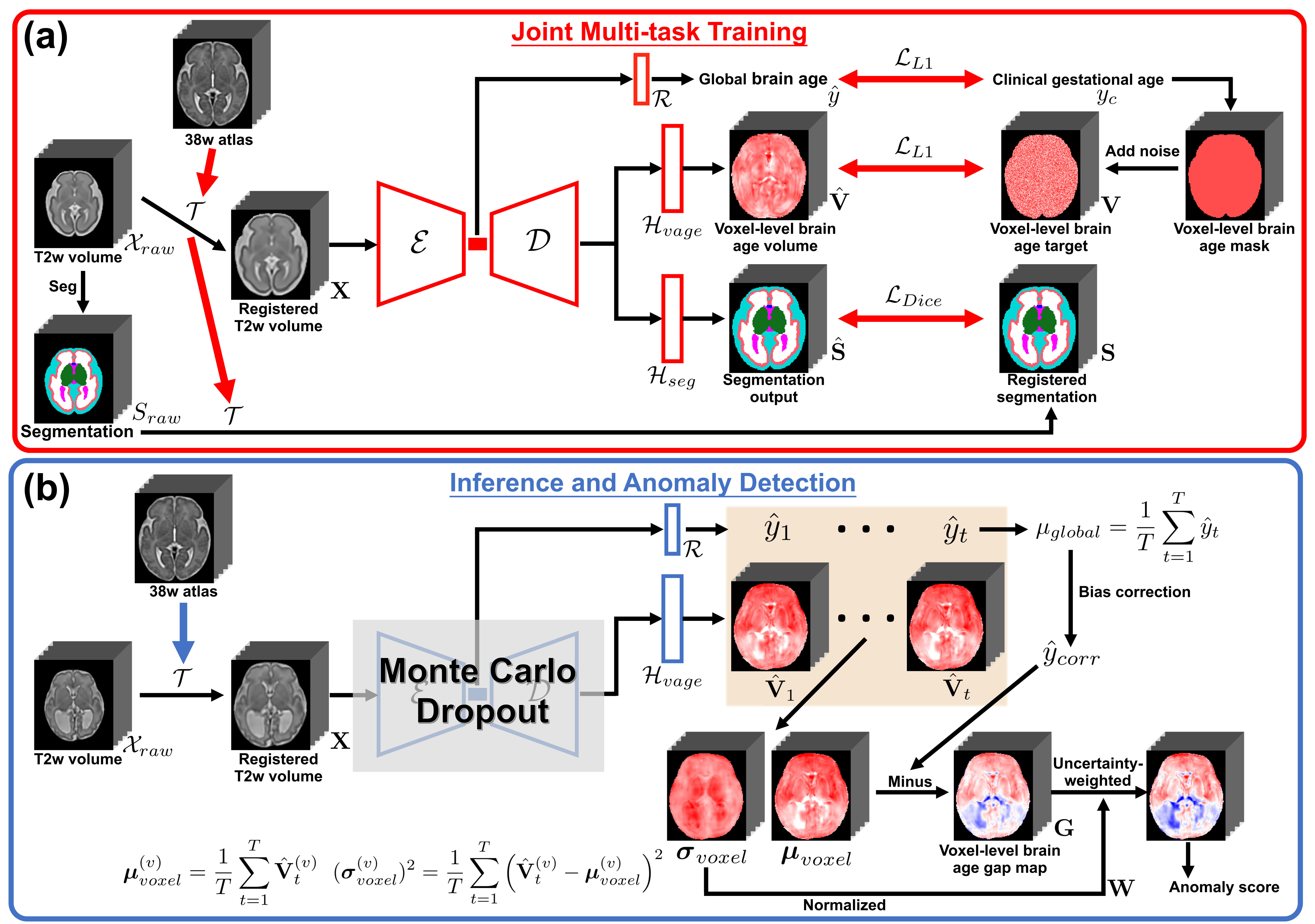}
    \caption{Overview of FetalBUAD. (a) Joint multi-task training: A shared encoder-decoder network predicts global brain age, voxel-level brain age, and brain segmentation from normal fetal brain MRIs. Noise perturbation of voxel-level targets encourages local morphological learning, while segmentation provides anatomical constraints. (b) Inference and anomaly detection: Monte Carlo Dropout estimates predictive means and epistemic uncertainty for global and voxel-level age predictions. The bias-corrected global age serves as the biological reference, and the voxel-level brain age gap is weighted by normalized uncertainty to generate the final anomaly map.}
    \label{fig1}
\end{figure*}

\section{related work}
Existing fetal brain UAD methods can be broadly categorized into two paradigms: brain age prediction-based methods \cite{single_slice,multi_slice,whole_brain,PANDA} and reconstruction-based methods \cite{10540605,AE,DAE,sftp,iter}. However, neither fully addresses the need for interpretable developmental assessment and spatially precise lesion localization, especially for subtle and anatomically heterogeneous abnormalities.

\subsection{Brain age prediction-based methods}
Brain age prediction-based methods model normative neurodevelopment and identify abnormal cases through discrepancies between predicted brain age and clinically recorded GA. Shi et al. introduced an attention-based deep ensemble for fetal brain age estimation from routine T2-weighted MRI, using predictive uncertainty and estimation confidence as anomaly markers \cite{single_slice}. Subsequent work applied a 2D single-channel CNN to multiplanar fetal MRI sections to detect ventriculomegaly and associated CNS abnormalities using predicted age differences \cite{multi_slice}. A lightweight 3D network enabled whole-brain age estimation and highlighted the importance of bias correction in brain-age delta estimation \cite{whole_brain}. More recently, PANDA extended age prediction to local image patches and used the maximum absolute age difference across patches as an anomaly biomarker, enabling coarse lesion localization \cite{PANDA}.
However, existing methods mainly produce case-level scores and cannot directly localize abnormalities at the voxel level. Their anomaly scores are also affected by noise in clinically recorded GA labels.

\subsection{Reconstruction-based methods}
Reconstruction-based methods identify anomalies through image reconstruction or feature-representation discrepancies. Standard 3D autoencoders and denoising autoencoders detect abnormal regions using pixel-wise reconstruction errors from corrupted inputs \cite{AE,DAE}. IterMask3D iteratively refines spatial masks towards anomalies using reconstruction errors and high-frequency structural guidance to reduce false positives in normal regions \cite{iter}. Feature-level teacher-student methods instead detect anomalies through discrepancies between teacher and student representations. Skip-TS uses heterogeneous encoder-decoder reverse distillation with skip connections and multiscale anomaly consistency \cite{10540605}, while 3D student-teacher feature pyramid matching applies self-supervised patch learning, ACS convolutions, and multiscale feature matching to volumetric MRI \cite{sftp}.

These methods generate spatial anomaly maps, but their evidence primarily reflects image restoration or feature-level discrepancies rather than developmental semantics. High-capacity reconstruction models may reproduce abnormal structures, while stronger corruption or masking can increase false positives in normal tissue. Moreover, their appearance-dependent representations are susceptible to domain shifts caused by fetal motion, contrast variations, imaging artifacts, and cross-site differences. Consequently, they localize appearance-level deviations but do not explicitly assess whether local fetal brain structures are developmentally appropriate.

\section{method}

\subsection{Data Preprocessing}
Let $\mathcal{X}_{raw}$ denote the T2-weighted fetal brain MRI volume. A publicly available fetal brain segmentation model \cite{TrustworthySeg} is employed to generate the corresponding brain segmentation $\mathcal{S}_{raw}$. Subsequently, an affine transformation $\mathcal{T}$ is performed to register $\mathcal{X}_{raw}$ to a 38-week fetal brain atlas volume \cite{atlas}. Specifically, this transformation is implemented using ANTsPy \cite{antspy} and involves only rigid rotation and isotropic scaling, without any nonlinear deformations. This alignment normalizes the global brain size to a unified and consistent scale, thereby compelling the network to prioritize intrinsic morphological patterns over absolute brain volume when predicting brain age. The corresponding brain segmentation is registered using the identical transformation $\mathcal{T}$ with nearest-neighbor interpolation to serve as a training target $\mathbf{S} = \mathcal{T}(S_{raw})$. Finally, $\mathcal{X}_{raw}$ is further normalized to zero mean and unit variance to serve as the final network input $\mathbf{X} =\mathcal{N}( \mathcal{T}(\mathcal{X}_{raw}))$.

\subsection{Joint Multi-task Network Training}
\subsubsection{Network Architecture} A multi-task framework is proposed to simultaneously predict the global brain age, the voxel-level brain age volume, and the brain segmentation result. Specifically, the network architecture consists of a shared encoder $\mathcal{E}(\cdot)$, a global regression head $\mathcal{R}(\cdot)$, a dual-purpose decoder $\mathcal{D}(\cdot)$, a voxel-level estimation head $\mathcal{H}_{vage}(\cdot)$, and a segmentation head $\mathcal{H}_{seg}(\cdot)$.

Given the input $\mathbf{X}$, the encoder first extracts a compact latent bottleneck representation $\mathbf{z} = \mathcal{E}(\mathbf{X})$ through dropout-regularized layers, which is subsequently processed via two parallel pathways: (1) Global Age Prediction Path: The global regression head $\mathcal{R}$ maps $\mathbf{z}$ to a scalar global brain age prediction, denoted as $\hat{y} = \mathcal{R}(\mathbf{z})$. (2) Dense Voxel-wise Path: The decoder $\mathcal{D}$ upsamples $\mathbf{z}$ to reconstruct high-resolution feature maps, which bifurcate into two convolutional heads: the voxel-level estimation head predicts a voxel-level brain age volume $\hat{\mathbf{V}}=\mathcal{H}_{vage}(\mathcal{D}(\mathbf{z}))$, while the segmentation head predicts the brain segmentation result $\hat{\mathbf{S}}=\mathcal{H}_{seg}(\mathcal{D}(\mathbf{z}))$. 

\subsubsection{Training Objective} For the input $\mathbf{X}$ with clinically recorded GA $y_{c}$, the voxel-level brain age regression target $\mathbf{V}$ is formulated as:
\begin{equation}
    \mathbf{V}^{(v)} = (y_{c} + \epsilon^{(v)}) \cdot \mathbb{M}_{\mathbf{S}}^{(v)}, \quad \text{where } \epsilon^{(v)} \sim \mathcal{U}(-\delta, \delta),
    \label{eq:noise}
\end{equation}
$\epsilon^{(v)}$ is a random noise term sampled independently for each voxel $v$ from a uniform distribution within a small range. $\mathbb{M}_{\mathbf{S}}$ denotes the binary brain mask derived from the registered segmentation $\mathbf{S}$. This perturbation strategy introduces subtle variations, compelling the network to capture fine-grained local texture variations rather than converging to a trivial solution \cite{VBAP}. Constraining the noise magnitude $\delta$ within a small range ensures that the perturbation does not disrupt the global developmental trend while effectively preventing identical gradient flows. 

The network is optimized by minimizing a composite loss function:
\begin{equation}
    \mathcal{L}_{total} = \mathcal{L}_{L1}(\hat{y}, y_{c}) + 
    \lambda_{vage} \mathcal{L}_{L1}(\hat{\mathbf{V}}, \mathbf{V})+
    \lambda_{seg} \mathcal{L}_{Dice}(\hat{\mathbf{S}}, \mathbf{S}), 
\end{equation}
where $\mathcal{L}_{L1}$ denotes the L1 loss, $\mathcal{L}_{Dice}$ is the Dice loss, and $\lambda_{vage}$ and $\lambda_{seg}$ are hyperparameters to scale the losses to the same order of magnitude and balance their contributions. Incorporating the segmentation task serves as an anatomical constraint, compelling the shared encoder to learn spatially consistent representations.

\subsubsection{Robustness to Noisy Clinically Recorded GA}
Clinical GA labels $y_{\text{c}}$ may contain measurement noise and systematic errors. Let $y^*$ denote the true biological GA, with
\begin{equation}
y_{\text{c}} = y^* + \eta, \quad
\mathbb{E}[\eta \mid \mathbf{X}] \approx 0,
\end{equation}
where $\eta$ represents measurement noise. The global brain age predictor $\hat{y} = \mathcal{R}(\mathcal{E}(\mathbf{X}))$ is trained using L1 loss, whose optimal predictor estimates the conditional median of clinical GA:
\begin{equation}
\hat{y}^* = \underset{\mathcal{R}\circ\mathcal{E}}{\arg\min}\;
\mathbb{E}_{\mathbf{X},y_{\text{c}}}
\left[\left|\mathcal{R}(\mathcal{E}(\mathbf{X}))-y_{\text{c}}\right|\right]
\approx \operatorname{Median}(y_{\text{c}}\mid\mathbf{X}).
\end{equation}
Under approximately symmetric measurement noise, this estimate can capture the underlying biological GA while being robust to variability in clinical labels.

\subsection{Inference and Anomaly Detection}
\subsubsection{Uncertainty Estimation via Monte Carlo Dropout}
To enable robust anomaly detection, we approximate a Bayesian neural network using MC Dropout during the inference phase. By keeping dropout active, $T$ stochastic forward passes are performed for a given input $\mathbf{X}$. The predictive mean and epistemic uncertainty are then derived for both the global and the voxel-level brain age estimation. For the global brain age, the aggregated mean $\mu_{global}$ and its corresponding variance $\sigma_{global}^2$ are given by:
\begin{equation}
    \mu_{global} = \frac{1}{T} \sum_{t=1}^{T} \hat{y}_{t}, \quad \sigma_{global}^2 = \frac{1}{T} \sum_{t=1}^{T} \left( \hat{y}_{t} - \mu_{global} \right)^2.
    \label{eq:uncertainty_glo}
\end{equation}
Similarly, for the voxel-level brain age, the mean age map $\boldsymbol{\mu}_{voxel}$ and the uncertainty map $\boldsymbol{\sigma}_{voxel}$ are computed as:
\begin{equation}
    \boldsymbol{\mu}_{voxel}^{(v)} = \frac{1}{T} \sum_{t=1}^{T} \hat{\mathbf{V}}_{t}^{(v)}, \quad (\boldsymbol{\sigma}_{voxel}^{(v)})^2 = \frac{1}{T} \sum_{t=1}^{T} \left( \hat{\mathbf{V}}_{t}^{(v)} - \boldsymbol{\mu}_{voxel}^{(v)} \right)^2,
    \label{eq:uncertainty_vox}
\end{equation}
where $\hat{y}_{t}$ and $\hat{\mathbf{V}}_{t}$ denote the scalar global brain age and voxel-level brain age volume output at the $t$-th stochastic pass under active dropout, respectively.

\subsubsection{Bias Correction}
Owing to model regularization and non-Gaussian distribution of GAs in the dataset \cite{whole_brain}, the network predictions exhibit the expected regression dilution, manifesting as a spurious correlation between the fetal brain age gap and the biological GA. 

The predicted mean global brain age is adjusted using a linear bias correction strategy \cite{Biascorr,UnetvolBAP}. Specifically, a linear model $\mu_{global} = \alpha y_{c} + \beta$ is fitted on the internal validation set. During inference on unseen data, the corrected global age $\hat{y}_{corr}$ is computed using the fixed parameters $\alpha$ and $\beta$: $\hat{y}_{corr} = (\mu_{global} - \beta)/\alpha$.

The bias correction is applied exclusively to the global age prediction, leaving the voxel-level age maps unadjusted. The proposed target randomization strategy (Eq.~\ref{eq:noise}) inherently discourages the voxel-level regressor from collapsing to the mean, thereby implicitly mitigating the regression dilution at the voxel-level.

\subsubsection{Anomaly Detection Metric}
The voxel-level brain age gap map $\mathbf{G}$ is defined as the voxel-wise difference of the mean voxel-level brain age volume and the corrected global brain age: $\mathbf{G}^{(v)} = \boldsymbol{\mu}_{voxel}^{(v)} - \hat{y}_{corr}$. This strategy ensures that $\mathbf{G}$ reflects purely structural deviations. Positive values in $\mathbf{G}$ imply advanced regional maturation, while negative values suggest developmental delays. 

Pathological regions typically exhibit higher epistemic uncertainty due to their deviation from the normal training manifold. To leverage this, we define an uncertainty-weighted deviation index that amplifies the magnitude of $\mathbf{G}$ based on the local uncertainty, and the final anomaly score is defined as the dispersion of the positive amplified deviations:
\begin{equation}
    \text{Score} = \text{Std}\left( \{ \mathbf{G}^{(v)} \cdot \mathbf{W}^{(v)} \mid \mathbf{G}^{(v)} \cdot \mathbf{W}^{(v)} > 0 \} \right),
\end{equation}
where $\mathbf{W}^{(v)} = 1 + (\boldsymbol{\sigma}_{voxel}^{(v)} - \min(\boldsymbol{\sigma}_{voxel})) / (\max(\boldsymbol{\sigma}_{voxel}) - \min(\boldsymbol{\sigma}_{voxel}))$ is a normalized weight factor mapped to the range [1, 2] that adaptively emphasizes high-uncertainty voxels during inference. Only positive deviations are retained, as structural anomalies tend to cause underestimation of the global brain age, leading pathological cases to exhibit significant positive gaps relative to normal cases.

\section{Dataset and implementation}

\renewcommand{\arraystretch}{1}

\begin{table*}[!ht]
\centering
\setlength{\tabcolsep}{2.4pt} 
\fontsize{8}{9.6}\selectfont   
\caption{Comparison of case-level anomaly detection performance on the internal test set and external test set. AUROC and AUPR are reported together with sensitivity at a fixed specificity of 0.80 and specificity at a fixed sensitivity of 0.80. Symbols indicate method categories: $^{\dagger}$ brain age prediction-based methods; $^{\ddagger}$ reconstruction-based methods. Best in \colorbox{blue!18}{blue}, runner-up in \colorbox{green!20}{green}.}
\label{tab:internal}
\newcommand{\mAge}[1]{#1$^{\dagger}$}
\newcommand{\mRec}[1]{#1$^{\ddagger}$}
\begin{tabular}{c|lcccc}
\toprule
 & Method & AUROC & AUPR & Sensitivity & Specificity \\
\midrule

\multirow{9}{*}{\begin{tabular}{c} \textbf{Internal Test Dataset:} \\ \textbf{Abnormal ($n=$~110)} \\ \textbf{\& Normal ($n=$~606)}  \end{tabular}}
& \mAge{Single-slice AAD}\cite{single_slice} & 0.656 [0.595, 0.713] & 0.340 [0.254, 0.432] & 0.436 [0.333, 0.532] & 0.340 [0.257, 0.412] \\
& \mAge{Multi-slice AAD}\cite{multi_slice}  & 0.596 [0.528, 0.657] & 0.280 [0.208, 0.364] & 0.355 [0.272, 0.450] & 0.188 [0.102, 0.398] \\
& \mAge{Whole-brain AAD}\cite{whole_brain}  & 0.584 [0.522, 0.643] & 0.269 [0.193, 0.350] & 0.318 [0.213, 0.420] & 0.252 [0.209, 0.342] \\
& \mAge{PANDA}\cite{PANDA}                        & 0.726 [0.674, 0.781] & 0.441 [0.355, 0.533] & 0.564 [0.467, 0.656] & 0.465 [0.346, 0.611] \\
& \mRec{3D AE}\cite{AE}              & \second{0.740} [0.688, 0.789] & 0.392 [0.297, 0.482] & 0.509 [0.409, 0.621] & 0.502 [0.367, 0.666] \\
& \mRec{3D DAE}\cite{DAE}              & 0.630 [0.571, 0.690] & 0.285 [0.210, 0.366] & 0.400 [0.287, 0.496] & 0.304 [0.213, 0.443] \\
& \mRec{3D STFPM}\cite{sftp}            & 0.665 [0.608, 0.717] & 0.291 [0.216, 0.375] & 0.418 [0.303, 0.510] & 0.422 [0.354, 0.492] \\
& \mRec{IterMask3D}\cite{iter}          & \best{0.905} [0.872, 0.937] & \best{0.779} [0.715, 0.840] & \second{0.800} [0.714, 0.876] & \second{0.787} [0.731, 0.931] \\
\rowcolor{gray!10} \cellcolor{white} & \textbf{\texttt{FetalBUAD}}  & \best{0.905} [0.870, 0.935] & \second{0.764} [0.694, 0.828] & \best{0.845} [0.750, 0.907] & \best{0.868} [0.715, 0.919] \\
\midrule

\multirow{9}{*}{\begin{tabular}{c} \textbf{Subgroup: ACC ($n=$~23)} \\ \textbf{\& Normal ($n=$~606)}\end{tabular}}
& \mAge{Single-slice AAD}\cite{single_slice} & 0.685 [0.559, 0.806] & 0.230 [0.074, 0.430] & 0.435 [0.210, 0.647] & 0.401 [0.075, 0.622] \\
& \mAge{Multi-slice AAD}\cite{multi_slice}  & 0.653 [0.534, 0.762] & 0.112 [0.039, 0.298] & 0.391 [0.192, 0.609] & 0.404 [0.138, 0.608] \\
& \mAge{Whole-brain AAD}\cite{whole_brain}  & 0.591 [0.451, 0.731] & 0.119 [0.033, 0.258] & 0.304 [0.118, 0.546] & 0.251 [0.059, 0.443] \\
& \mAge{PANDA}\cite{PANDA}                        & 0.754 [0.639, 0.857] & 0.255 [0.095, 0.440] & 0.565 [0.350, 0.765] & 0.528 [0.285, 0.752] \\
& \mRec{3D AE}\cite{AE}               & 0.751 [0.645, 0.839] & 0.101 [0.050, 0.212] & 0.478 [0.269, 0.720] & 0.502 [0.368, 0.768] \\
& \mRec{3D DAE}\cite{DAE}              & 0.613 [0.467, 0.757] & 0.088 [0.037, 0.218] & 0.435 [0.222, 0.650] & 0.224 [0.046, 0.600] \\
& \mRec{3D STFPM}\cite{sftp}            & 0.688 [0.589, 0.777] & 0.086 [0.039, 0.222] & 0.348 [0.160, 0.571] & 0.465 [0.268, 0.647] \\
& \mRec{IterMask3D}\cite{iter}          & \second{0.920} [0.842, 0.982] & \best{0.763} [0.588, 0.907] & \second{0.826} [0.667, 0.963] & \best{0.949} [0.479, 1.000] \\
\rowcolor{gray!10} \cellcolor{white} & \textbf{\texttt{FetalBUAD}}  & \best{0.949} [0.881, 0.991] & \second{0.753} [0.575, 0.896] & \best{0.913} [0.769, 1.000] & \second{0.939} [0.629, 0.997] \\
\midrule

\multirow{9}{*}{\begin{tabular}{c} \textbf{Subgroup: DWM ($n=$~22)} \\ \textbf{\& Normal ($n=$~606)}  \end{tabular}}
& \mAge{Single-slice AAD}\cite{single_slice} & 0.701 [0.577, 0.823] & 0.110 [0.045, 0.286] & 0.500 [0.273, 0.714] & 0.378 [0.280, 0.679] \\
& \mAge{Multi-slice AAD}\cite{multi_slice}  & 0.576 [0.420, 0.717] & 0.161 [0.038, 0.327] & 0.364 [0.187, 0.591] & 0.188 [0.066, 0.484] \\
& \mAge{Whole-brain AAD}\cite{whole_brain}  & 0.638 [0.507, 0.766] & 0.237 [0.081, 0.416] & 0.364 [0.174, 0.588] & 0.403 [0.181, 0.538] \\
& \mAge{PANDA}\cite{PANDA}                        & 0.765 [0.658, 0.867] & 0.305 [0.115, 0.500] & 0.636 [0.428, 0.833] & 0.474 [0.312, 0.851] \\
& \mRec{3D AE}\cite{AE}               & 0.886 [0.813, 0.940] & 0.412 [0.205, 0.608] & 0.773 [0.600, 0.950] & 0.787 [0.650, 0.903] \\
& \mRec{3D DAE}\cite{DAE}              & 0.643 [0.527, 0.748] & 0.096 [0.029, 0.213] & 0.318 [0.136, 0.533] & 0.449 [0.185, 0.619] \\
& \mRec{3D STFPM}\cite{sftp}            & 0.729 [0.615, 0.828] & 0.203 [0.060, 0.375] & 0.455 [0.250, 0.667] & 0.531 [0.258, 0.638] \\
& \mRec{IterMask3D}\cite{iter}          & \second{0.920} [0.867, 0.971] & \second{0.610} [0.408, 0.792] & \second{0.818} [0.625, 0.963] & \second{0.835} [0.658, 0.977] \\
\rowcolor{gray!10} \cellcolor{white} & \textbf{\texttt{FetalBUAD}}  & \best{0.946} [0.898, 0.983] & \best{0.687} [0.487, 0.850] & \best{0.909} [0.769, 1.000] & \best{0.913} [0.740, 0.997] \\
\midrule

\multirow{9}{*}{\begin{tabular}{c} \textbf{Subgroup: GMH-IVH ($n=$~65)} \\ \textbf{\& Normal ($n=$~606)}  \end{tabular}}
& \mAge{Single-slice AAD}\cite{single_slice} & 0.630 [0.557, 0.704] & 0.214 [0.135, 0.316] & 0.415 [0.279, 0.529] & 0.272 [0.221, 0.395] \\
& \mAge{Multi-slice AAD}\cite{multi_slice}  & 0.583 [0.499, 0.663] & 0.172 [0.106, 0.258] & 0.338 [0.213, 0.449] & 0.114 [0.064, 0.423] \\
& \mAge{Whole-brain AAD}\cite{whole_brain}  & 0.563 [0.491, 0.638] & 0.144 [0.091, 0.212] & 0.308 [0.169, 0.419] & 0.252 [0.210, 0.333] \\
& \mAge{PANDA}\cite{PANDA}                        & 0.703 [0.623, 0.774] & 0.318 [0.202, 0.431] & 0.538 [0.394, 0.649] & 0.391 [0.231, 0.616] \\
& \mRec{3D AE}\cite{AE}               & 0.686 [0.622, 0.752] & 0.211 [0.141, 0.306] & 0.431 [0.319, 0.574] & 0.378 [0.311, 0.635] \\
& \mRec{3D DAE}\cite{DAE}              & 0.632 [0.565, 0.705] & 0.212 [0.134, 0.305] & 0.415 [0.262, 0.533] & 0.305 [0.218, 0.453] \\
& \mRec{3D STFPM}\cite{sftp}           & 0.636 [0.570, 0.704] & 0.159 [0.112, 0.235] & 0.431 [0.279, 0.542] & 0.380 [0.317, 0.461] \\
& \mRec{IterMask3D}\cite{iter}          & \best{0.895} [0.854, 0.936] & \best{0.695} [0.594, 0.793] & \second{0.785} [0.662, 0.897] & \best{0.779} [0.656, 0.914] \\
\rowcolor{gray!10} \cellcolor{white} & \textbf{\texttt{FetalBUAD}}  & \second{0.876} [0.830, 0.921] & \second{0.618} [0.493, 0.716] & \best{0.800} [0.667, 0.885] & \second{0.741} [0.641, 0.894] \\
\midrule

\multirow{9}{*}{\begin{tabular}{c} \textbf{Dataset: External Test} \\ \textbf{Abnormal ($n=$~49)} \\ \textbf{\& Normal ($n=$~31)}  \end{tabular}}
& \mAge{Single-slice AAD}\cite{single_slice} & 0.512 [0.380, 0.639] & 0.680 [0.540, 0.808] & 0.286 [0.157, 0.449] & 0.226 [0.000, 0.444] \\
& \mAge{Multi-slice AAD}\cite{multi_slice}  & 0.609 [0.477, 0.723] & 0.704 [0.558, 0.843] & 0.327 [0.130, 0.660] & 0.226 [0.083, 0.539] \\
& \mAge{Whole-brain AAD}\cite{whole_brain}& 0.474 [0.348, 0.616] & 0.588 [0.465, 0.762] & 0.224 [0.000, 0.531] & 0.097 [0.000, 0.281] \\
& \mAge{PANDA}\cite{PANDA}                        & 0.461 [0.340, 0.601] & 0.572 [0.443, 0.732] & 0.224 [0.000, 0.422] & 0.129 [0.000, 0.353] \\
& \mRec{3D AE}\cite{AE}               & 0.708 [0.582, 0.826] & 0.749 [0.600, 0.893] & 0.531 [0.127, 0.769] & 0.516 [0.189, 0.778] \\
& \mRec{3D DAE}\cite{DAE}              & 0.554 [0.433, 0.676] & 0.702 [0.568, 0.825] & 0.347 [0.159, 0.500] & 0.161 [0.032, 0.434] \\
& \mRec{3D STFPM}\cite{sftp}            & \second{0.766} [0.653, 0.873] & \second{0.833} [0.720, 0.922] & \second{0.571} [0.326, 0.813] & \second{0.548} [0.364, 0.800] \\
& \mRec{IterMask3D}\cite{iter}          & 0.711 [0.597, 0.818] & 0.815 [0.698, 0.906] & 0.551 [0.280, 0.741] & 0.387 [0.192, 0.714] \\
\rowcolor{gray!10} \cellcolor{white} & \textbf{\texttt{FetalBUAD}}              & \best{0.800} [0.697, 0.885] & \best{0.874} [0.786, 0.937] & \best{0.673} [0.470, 0.824] & \best{0.613} [0.371, 0.844] \\
\bottomrule
\end{tabular}
\end{table*}

\begin{figure*}[!h]
	\centering
	\includegraphics[width=0.98\linewidth]{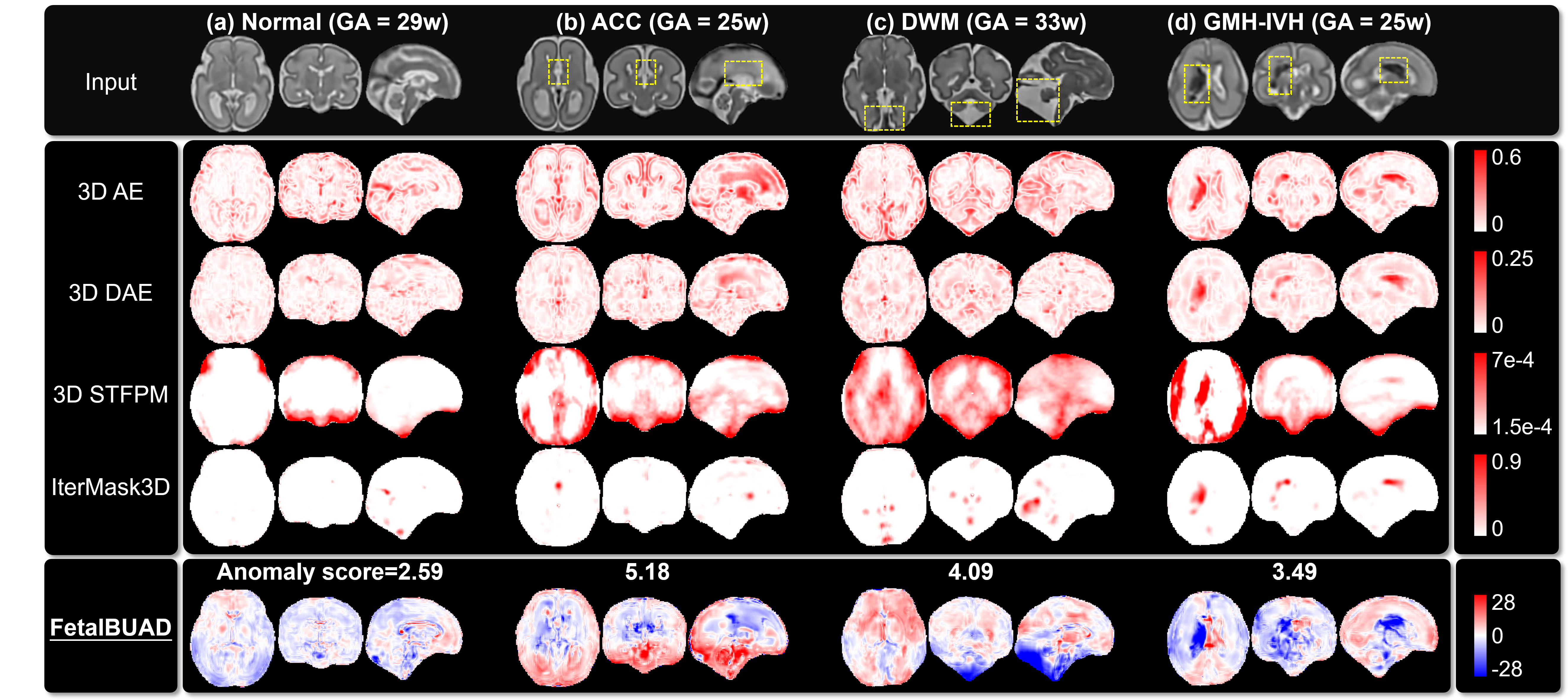}
	\caption{Lesion localization in representative normal and pathological cases from the internal test set. Yellow dashed boxes indicate ground-truth lesions. Pathological findings include corpus callosum agenesis in ACC, posterior fossa cystic dilation and cerebellar hypoplasia in DWM, and hypointense hemorrhagic regions in GMH-IVH. Rows show anomaly maps from reconstruction-based baselines (3D AE, 3D DAE, 3D STFPM, and IterMask3D) and the proposed FetalBUAD.}
    \label{fig2}
\end{figure*}

\subsection{Datasets}
\subsubsection{Training and internal test datasets}
T2-weighted (T2w) thick-slice fetal brain MRI scans were retrospectively collected from 1,400 pregnant women (GA: 20--39 weeks) at a single center under IRB approval using three scanners (1.5T and 3T). 3D brain volumes were reconstructed using NiftyMIC \cite{niftymic,LIU2026103994}. Following clinician quality assessment, cases with severe motion, incomplete brain coverage, or other compromising artifacts were excluded, leaving 1,317 cases: 1,207 normal and 110 pathological, including agenesis of the corpus callosum (ACC, $n=$~23), Dandy-Walker malformation (DWM, $n=$~22), and germinal matrix and intraventricular hemorrhage (GMH-IVH, $n=$~65). Of the normal cases, 487 were used for training, 114 for validation, and the remaining 606, together with all 110 pathological cases, for internal testing.

\subsubsection{External test dataset}
The FeTA dataset \cite{FeTA1,FeTA2}, from the 2024 Fetal Tissue Annotation and Segmentation Challenge, was used for external testing to assess cross-domain generalization. It comprised 3D fetal brain MRI volumes from 31 normal and 49 pathological fetuses (GA: 20--35 weeks).

\subsection{Implementation Details and Comparison Methods}
\subsubsection{Implementation Details} Clinical GA labels were derived from first-trimester ultrasound crown-rump length measurements and expressed in days. The network used a four-level 3D U-Net with channels [32, 64, 128, 256]. A global age head branched from the encoder bottleneck, comprising adaptive average pooling, dropout (0.1), and a fully connected layer. Two separate convolutional heads applied to the decoder output predicted voxel-level brain age and nine-class segmentation. Additional dropout (0.3) was applied at the bottleneck. Hyperparameter $\delta=2$ days, $T=36$, $\lambda_{vage}=1$ and $\lambda_{seg}=1000$ were chosen during early training to scale the 3 losses to the same order of magnitude. Training was performed using the Adam optimizer with a cosine annealing scheduler (initial learning rate: 0.001), over 150 epochs with batch size 8 on one NVIDIA A800 GPU.

\subsubsection{Comparison Methods} Brain age prediction-based methods, which use the absolute age difference (AAD) between predicted GA and clinically recorded GA, include Single-slice AAD \cite{single_slice}, Multi-slice AAD \cite{multi_slice}, Whole-brain AAD \cite{whole_brain}, and Patch-based method PANDA \cite{PANDA}. Reconstruction-based methods include 3D AE \cite{AE}, 3D DAE \cite{DAE}, 3D STFPM \cite{sftp}, and IterMask3D \cite{iter}. All comparison methods were implemented following their original training and inference protocols, and evaluated under exactly matched settings.

\section{Results and Analysis}

\begin{table*}[!ht]
\centering
\setlength{\tabcolsep}{3.5pt} 
\fontsize{8}{9.6}\selectfont  
\caption{Ablation study on the internal test set. AUROC and AUPR are reported together with sensitivity at a fixed specificity of 0.80 and specificity at a fixed sensitivity of 0.80. Best in \colorbox{blue!18}{blue}, runner-up in \colorbox{green!20}{green}.}
\label{tab:ablation}
\begin{tabular}{c|lcccc}
\toprule
 & Method & AUROC & AUPR & Sensitivity & Specificity \\
\midrule
\multirow{5}{*}{\begin{tabular}{c} \textbf{Internal Test Dataset:} \\ \textbf{Abnormal ($n=$~110)} \\ \textbf{\& Normal ($n=$~606)} \end{tabular}}
& w/o registration to 38w & 0.870 [0.827, 0.911] & 0.718 [0.638, 0.794] & \second{0.809} [0.720, 0.877] & \second{0.802} [0.662, 0.878] \\
& w/o segmentation task   & 0.829 [0.786, 0.870] & 0.616 [0.529, 0.697] & 0.655 [0.561, 0.762] & 0.652 [0.503, 0.779] \\
& w/o uncertainty-weighted  & \second{0.887} [0.849, 0.920] & \second{0.737} [0.665, 0.804] & 0.800 [0.722, 0.868] & 0.759 [0.625, 0.887] \\
& Metric: mean            & 0.881 [0.843, 0.914] & 0.711 [0.636, 0.778] & 0.755 [0.675, 0.841] & 0.748 [0.643, 0.862] \\
\rowcolor{gray!10} \cellcolor{white} & \textbf{\texttt{FetalBUAD}} & \best{0.905} [0.870, 0.935] & \best{0.764} [0.694, 0.828] & \best{0.845} [0.750, 0.907] & \best{0.868} [0.715, 0.919] \\
\bottomrule
\end{tabular}
\end{table*}

\subsection{Lesion Localization Performance}
As illustrated in Fig.~\ref{fig2}, the voxel-level brain age gap maps generated by \textbf{\texttt{FetalBUAD}} remain close to zero in normal brains, indicating that the biological brain age reference effectively suppresses normal developmental variations and produces stable localization maps. For pathological subjects, the proposed method highlights anatomically meaningful abnormalities across lesion types, including the absent corpus callosum in ACC, the enlarged posterior fossa and cerebellar hypoplasia in DWM, and focal hemorrhagic regions in GMH-IVH.
Compared with reconstruction-based approaches, which often produce diffuse responses or scattered false positives in normal tissues, \textbf{\texttt{FetalBUAD}} generates sharper, more anatomically confined anomaly maps. Its responses closely correspond to clinician-annotated lesion locations, suggesting that voxel-level brain age estimation captures localized developmental deviations rather than reconstruction residuals. These results demonstrate improved spatial specificity and lesion boundary delineation across heterogeneous fetal brain abnormalities.

\subsection{Anomaly Detection Performance}
\subsubsection{Internal Test}
As summarized in Table~\ref{tab:internal}, \textbf{\texttt{FetalBUAD}} achieves an AUROC of 0.905, matching the best-performing reconstruction-based method IterMask3D while outperforming all brain-age prediction baselines and the remaining reconstruction-based methods. Beyond AUROC, \textbf{\texttt{FetalBUAD}} achieves the highest sensitivity (0.845) and specificity (0.868) at the predefined operating thresholds, indicating a favorable balance between pathological case detection and false-positive control.

Disease-specific evaluation shows that \textbf{\texttt{FetalBUAD}} achieves the best performance on ACC (AUROC 0.949) and DWM (AUROC 0.946), outperforming IterMask3D (0.920 for both). These disorders primarily involve localized structural malformations rather than prominent intensity changes, highlighting the advantage of voxel-level brain age estimation for detecting regional developmental abnormalities. Conversely, IterMask3D performs slightly better on GMH-IVH (0.895 vs.\ 0.876), potentially because hemorrhagic lesions exhibit conspicuous intensity changes that reconstruction-based methods can readily capture.

\subsubsection{External Test}
As shown in Table~\ref{tab:internal}, \textbf{\texttt{FetalBUAD}} achieves the best performance across all metrics on the external FeTA dataset without target-domain fine-tuning, including AUROC (0.800), AUPR (0.874), sensitivity (0.673), and specificity (0.613). These results suggest that the proposed biological brain age representation generalizes more effectively across different acquisition protocols and institutions than reconstruction-based anomaly modeling. By assessing developmental consistency rather than reconstruction fidelity, the method may be less sensitive to domain shifts, improving cross-dataset robustness.

\subsection{Ablation Study}
We conducted ablation experiments on the internal test set to validate the contributions of key components (Table~\ref{tab:ablation}).

\subsubsection{\textbf{Removing affine registration to the 38-week atlas}} AUROC, AUPR, Sensitivity, and Specificity declined by 3.9\%, 6.0\%, 4.3\%, and 7.6\%, respectively. This suggests that original-sized brain volumes cause the model to rely disproportionately on brain size rather than internal structural relationships for GA prediction, degrading anomaly detection.

\subsubsection{\textbf{Excluding the auxiliary brain segmentation task in training}} This caused the largest performance drops, with AUROC, AUPR, Sensitivity, and Specificity decreasing by 8.4\%, 19.4\%, 22.5\%, and 24.9\%, respectively. This highlights the role of segmentation in providing anatomical constraints and promoting spatially consistent morphological representations.

\subsubsection{\textbf{Discarding uncertainty weighting during inference}} AUROC, AUPR, Sensitivity, and Specificity decreased by 2.0\%, 3.5\%, 5.3\%, and 12.6\%, respectively. These results confirm that uncertainty weighting emphasizes pathological regions in voxel-level brain age gap maps, improving detection.

\subsubsection{\textbf{Replacing standard deviation (std) with mean deviation for anomaly scoring}} AUROC, AUPR, Sensitivity, and Specificity dropped by 2.7\%, 6.9\%, 10.7\%, and 13.8\%, respectively. This indicates that standard deviation captures spatial heterogeneity in deviations and is a more sensitive anomaly indicator than the average shift alone.

\section{Conclusion}
We presented \textbf{\texttt{FetalBUAD}}, an uncertainty-aware framework for unsupervised fetal brain anomaly detection based on voxel-level biological brain age estimation. By establishing an internal biological age reference rather than relying on noisy clinical GA labels, \textbf{\texttt{FetalBUAD}} enables case-level anomaly detection and anatomically interpretable lesion localization. This work highlights voxel-level biological age modeling as a promising alternative to image consistency-based anomaly detection. Future work will extend the framework to broader fetal brain abnormalities and longitudinal MRI analysis.

\section{ACKNOWLEDGMENT}
This work was supported by the National Natural Science Foundation of China (grant number 82572198, 82672619), the Scientific Research Project of Sichuan Medical Association (grant number 2024HR130), the Science and Technology Department of Sichuan Province, China (grant number 25SYSX0255), the Tsinghua University Startup Fund, and the Tsinghua University Dushi Program (grant numbers 20241080026, 20251080056).

\bibliographystyle{IEEEtran}
\bibliography{IEEEabrv,reference.bib}{}

@article{10.1016/j.jksuci.2023.101647,
author = {Chowdhury, Anjir Ahmed and Hasan Mahmud, S.M. and Shahjalal Hoque, Khadija Kubra and Ahmed, Kawsar and Bui, Francis M. and Lio, Pietro and Moni, Mohammad Ali and Al-Zahrani, Fahad Ahmed},
title = {StackFBAs: Detection of fetal brain abnormalities using CNN with stacking strategy from MRI images},
year = {2023},
issue_date = {Sep 2023},
publisher = {Elsevier Science Inc.},
address = {USA},
volume = {35},
number = {8},
issn = {1319-1578},
journal = {Journal of King Saud University - Computer and Information Sciences},
month = sep,
numpages = {14}
}

@InProceedings{10.1007/978-3-032-05997-0_11,
author="Liu, Mingxuan
and Li, Yijin
and Zhu, Juncheng
and Yang, Hongjia
and Huang, Yiming
and Li, Haoxiang
and Chen, Yifei
and Bai, Xuguang
and Liao, Yi
and Qu, Haibo
and Tian, Qiyuan",
title="FetalExtract-LLM: Structured Information Extraction from Free-Text Fetal MRI Reports Based on Privacy-Ensuring Open-Weights Large Language Models",
booktitle="Perinatal, Preterm and Paediatric Image Analysis",
year="2026",
publisher="Springer Nature Switzerland",
address="Cham",
pages="119--129",
isbn="978-3-032-05997-0"
}

@article{PANDA,
    author = {Hao, Yingqi and Liu, Mingxuan and Zhu, Juncheng and Yang, Hongjia and Li, Haoxiang and Kang, Min and Song, Yan and Lai, Hua and Zhou, Xiaoling and Ning, Gang and Liao, Yi and Qu, Haibo and Tian, Qiyuan},
    title = {PANDA: Patch-based Unsupervised Deep Learning for Brain Anomaly Detection via Age Prediction in Fetal MRI},
    journal = {Imaging Neuroscience},
    year = {2026},
    month = {06},
    issn = {2837-6056},
    doi = {10.1162/IMAG.a.1293},
    url = {https://doi.org/10.1162/IMAG.a.1293},
    eprint = {https://direct.mit.edu/imag/article-pdf/doi/10.1162/IMAG.a.1293/2607082/imag.a.1293.pdf},
}

@article{AE,
title = {Unsupervised anomaly detection in brain MRI: Learning abstract distribution from massive healthy brains},
journal = {Computers in Biology and Medicine},
volume = {154},
pages = {106610},
year = {2023},
issn = {0010-4825},
author = {Guoting Luo and Wei Xie and Ronghui Gao and Tao Zheng and Lei Chen and Huaiqiang Sun}
}

@article{whole_brain,
title = {Accurate brain age prediction with lightweight deep neural networks},
journal = {Medical Image Analysis},
volume = {68},
pages = {101871},
year = {2021},
issn = {1361-8415},
author = {Han Peng and Weikang Gong and Christian F. Beckmann and Andrea Vedaldi and Stephen M. Smith}
}

@article{single_slice,
title = {Fetal brain age estimation and anomaly detection using attention-based deep ensembles with uncertainty},
journal = {NeuroImage},
volume = {223},
pages = {117316},
year = {2020},
issn = {1053-8119},
author = {Wen Shi and Guohui Yan and Yamin Li and Haotian Li and Tingting Liu and Cong Sun and Guangbin Wang and Yi Zhang and Yu Zou and Dan Wu}
}

@InProceedings{DAE,
  title = 	 {Denoising Autoencoders for Unsupervised Anomaly Detection in Brain MRI},
  author =       {Kascenas, Antanas and Pugeault, Nicolas and O'Neil, Alison Q.},
  booktitle = 	 {Proceedings of The 5th International Conference on Medical Imaging with Deep Learning},
  pages = 	 {653--664},
  year = 	 {2022},
  editor = 	 {Konukoglu, Ender and Menze, Bjoern and Venkataraman, Archana and Baumgartner, Christian and Dou, Qi and Albarqouni, Shadi},
  volume = 	 {172},
  series = 	 {Proceedings of Machine Learning Research},
  month = 	 {06--08 Jul},
  publisher =    {PMLR}
}

@article{multi_slice,
author = {Yun, Hyuk Jin and Lee, Han-Jui and You, Sungmin and Lee, Joo                         Young and Aguirre-Chavez, Jerjes and Vasung, Lana and Lee, Hyun                         Ju and Tarui, Tomo and Feldman, Henry                             A. and Grant, P.                         Ellen and Im, Kiho},
title = {Deep Learning–based Brain Age Prediction Using MRI to Identify                     Fetuses with Cerebral Ventriculomegaly},
journal = {Radiology: Artificial Intelligence},
volume = {7},
number = {2},
pages = {e240115},
year = {2025}
}

@InProceedings{sftp,
  title = 	 {A Patch-based Student-Teacher Pyramid Matching Approach to Anomaly Detection in 3D Magnetic Resonance Imaging},
  author =       {Schwarz, Johannes and Will, Lena and Wellmer, J\"org and Mosig, Axel},
  booktitle = 	 {Proceedings of The 7th International Conference on Medical Imaging with Deep Learning},
  pages = 	 {1357--1370},
  year = 	 {2024},
  editor = 	 {Burgos, Ninon and Petitjean, Caroline and Vakalopoulou, Maria and Christodoulidis, Stergios and Coupe, Pierrick and Delingette, Hervé and Lartizien, Carole and Mateus, Diana},
  volume = 	 {250},
  series = 	 {Proceedings of Machine Learning Research},
  month = 	 {03--05 Jul},
  publisher =    {PMLR}
}

@article{LIU2026103994,
title = {Quality-label-free Fetal Brain MRI Quality Control Based on Image Orientation Recognition Uncertainty},
journal = {Medical Image Analysis},
pages = {103994},
year = {2026},
issn = {1361-8415},
author = {Mingxuan Liu and Yi Liao and Haoxiang Li and Juncheng Zhu and Hongjia Yang and Yingqi Hao and Haibo Qu and Qiyuan Tian}
}

@ARTICLE{10540605,
   author={Liu, Mingxuan and Jiao, Yunrui and Lu, Jingqiao and Chen, Hong},
   journal={IEEE Transactions on Instrumentation and Measurement}, 
   title={Anomaly Detection for Medical Images Using Teacher–Student Model With Skip Connections and Multiscale Anomaly Consistency}, 
   year={2024},
   volume={73},
   number={},
   pages={1-15}}

@article{iter,
title = {IterMask3D: Unsupervised anomaly detection and segmentation with test-time iterative mask refinement in 3D brain MRI},
journal = {Medical Image Analysis},
volume = {107},
pages = {103763},
year = {2026},
issn = {1361-8415},
author = {Ziyun Liang and Xiaoqing Guo and Wentian Xu and Yasin Ibrahim and Natalie Voets and Pieter M. Pretorius and J. Alison Noble and Konstantinos Kamnitsas}
}

@ARTICLE{TrustworthySeg,
  author={Fidon, Lucas and Aertsen, Michael and Kofler, Florian and Bink, Andrea and David, Anna L. and Deprest, Thomas and Emam, Doaa and Guffens, Frédéric and Jakab, András and Kasprian, Gregor and Kienast, Patric and Melbourne, Andrew and Menze, Bjoern and Mufti, Nada and Pogledic, Ivana and Prayer, Daniela and Stuempflen, Marlene and Van Elslander, Esther and Ourselin, Sébastien and Deprest, Jan and Vercauteren, Tom},
  journal={IEEE Transactions on Pattern Analysis and Machine Intelligence}, 
  title={A Dempster-Shafer Approach to Trustworthy AI With Application to Fetal Brain MRI Segmentation}, 
  year={2024},
  volume={46},
  number={5},
  pages={3784-3795}}

@article{atlas,
  title={A normative spatiotemporal MRI atlas of the fetal brain for automatic segmentation and analysis of early brain growth},
  author={Gholipour, Ali and Rollins, Caitlin K and Velasco-Annis, Clemente and Ouaalam, Abdelhakim and Akhondi-Asl, Alireza and Afacan, Onur and Ortinau, Cynthia M and Clancy, Sean and Limperopoulos, Catherine and Yang, Edward and others},
  journal={Scientific reports},
  volume={7},
  number={1},
  pages={476},
  year={2017},
  publisher={Nature Publishing Group UK London}}

@article{Biascorr,
title = {Estimation of brain age delta from brain imaging},
journal = {NeuroImage},
volume = {200},
pages = {528-539},
year = {2019},
issn = {1053-8119},
author = {Stephen M. Smith and Diego Vidaurre and Fidel Alfaro-Almagro and Thomas E. Nichols and Karla L. Miller}
}

@ARTICLE{UnetvolBAP,
    
AUTHOR={Popescu, Sebastian G.  and Glocker, Ben  and Sharp, David J.  and Cole, James H. },
           
TITLE={Local Brain-Age: A U-Net Model},
          
JOURNAL={Frontiers in Aging Neuroscience},
          
VOLUME={Volume 13 - 2021},
  
YEAR={2021},
ISSN={1663-4365}}

@InProceedings{MCdropout,
  title = 	 {Dropout as a Bayesian Approximation: Representing Model Uncertainty in Deep Learning},
  author = 	 {Gal, Yarin and Ghahramani, Zoubin},
  booktitle = 	 {Proceedings of The 33rd International Conference on Machine Learning},
  pages = 	 {1050--1059},
  year = 	 {2016},
  editor = 	 {Balcan, Maria Florina and Weinberger, Kilian Q.},
  volume = 	 {48},
  series = 	 {Proceedings of Machine Learning Research},
  address = 	 {New York, New York, USA},
  month = 	 {20--22 Jun},
  publisher =    {PMLR}
}

@article{epilepsy,
author = {Urhoj, Stine Kjaer and Morris, Joan and Loane, Maria and Ballardini, Elisa and Barrachina-Bonet, Laia and Cavero-Carbonell, Clara and Coi, Alessio and Gissler, Mika and Given, Joanne and Heino, Anna and Jordan, Sue and Neville, Amanda and Santoro, Michele and Tan, Joachim and Tucker, David and Wellesley, Diana and Garne, Ester and Damkjaer, Mads},
title = {Higher risk of cerebral palsy, seizures/epilepsy, visual- and hearing impairments, cancer, injury and child abuse in children with congenital anomalies: Data from the EUROlinkCAT study},
journal = {Acta Paediatrica},
volume = {113},
number = {5},
pages = {1024-1031},
eprint = {https://onlinelibrary.wiley.com/doi/pdf/10.1111/apa.17136},
year = {2024}
}

@article{autism,
    author = {Ortug, Alpen and Guo, Yurui and Feldman, Henry A and Ou, Yangming and Warren, Jose Luis Alatorre and Dieuveuil, Harrison and Baumer, Nicole T and Faja, Susan K and Takahashi, Emi},
    title = {Autism-associated brain differences can be observed in utero using MRI},
    journal = {Cerebral Cortex},
    volume = {34},
    number = {4},
    pages = {bhae117},
    year = {2024},
    month = {04},
    issn = {1460-2199},
    eprint = {https://academic.oup.com/cercor/article-pdf/34/4/bhae117/57202231/bhae117.pdf},
}

@article{intellectual,
    author = {Russ, Jeffrey B and Agarwal, Sonika and Venkatesan, Charu and Scelsa, Barbara and Vollmer, Brigitte and Tarui, Tomo and Pardo, Andrea C and Lemmon, Monica E and Mulkey, Sarah B and Hart, Anthony R and Nagaraj, Usha D and Kuller, Jeffrey A and Whitehead, Matthew T and Cohen, Jennifer L and Gebb, Juliana S and Glenn, Orit A and Norton, Mary E and Gano, Dawn},
    title = {Fetal malformations of cortical development: review and clinical guidance},
    journal = {Brain},
    volume = {148},
    number = {6},
    pages = {1888-1903},
    year = {2025},
    month = {03},
    issn = {0006-8950}
}

@article{natureHB,
  title={Trajectories of brain and behaviour development in the womb, at birth and through infancy},
  author={Meredith Weiss, Staci and Aydin, Ezra and Lloyd-Fox, Sarah and Johnson, Mark H},
  journal={Nature Human Behaviour},
  volume={8},
  number={7},
  pages={1251--1262},
  year={2024},
  publisher={Nature Publishing Group UK London}
}

@article{fetalMRI,
title = {Use of MRI in the diagnosis of fetal brain abnormalities in utero (MERIDIAN): a multicentre, prospective cohort study},
journal = {The Lancet},
volume = {389},
number = {10068},
pages = {538-546},
year = {2017},
issn = {0140-6736},
author = {Paul D Griffiths and Michael Bradburn and Michael J Campbell and Cindy L Cooper and Ruth Graham and Deborah Jarvis and Mark D Kilby and Gerald Mason and Cara Mooney and Stephen C Robson and Allan Wailoo}
}

@article{VBAP,
    title = "A voxel-level approach to brain age prediction: A method to assess regional brain aging",
    author = "Gianchandani, Neha and Dibaji, Mahsa and Ospel, Johanna and Vega, Fernando and Bento, Mariana and MacDonald, M. Ethan and Souza, Roberto",
    journal = "Machine Learning for Biomedical Imaging",
    volume = "2",
    issue = "April 2024 issue",
    year = "2024",
    pages = "761--795",
    issn = "2766-905X"
}

@article{FeTA1,
  title={An automatic multi-tissue human fetal brain segmentation benchmark using the fetal tissue annotation dataset},
  author={Payette, Kelly and de Dumast, Priscille and Kebiri, Hamza and Ezhov, Ivan and Paetzold, Johannes C and Shit, Suprosanna and Iqbal, Asim and Khan, Romesa and Kottke, Raimund and Grehten, Patrice and others},
  journal={Scientific data},
  volume={8},
  number={1},
  pages={167},
  year={2021},
  publisher={Nature Publishing Group UK London}
}

@article{FeTA2,
  title = {Fetal brain tissue annotation and segmentation challenge results},
  journal = {Medical Image Analysis},
  volume = {88},
  pages = {102833},
  year = {2023},
  issn = {1361-8415},
  author={Payette, Kelly and Li, Hongwei Bran and de Dumast, Priscille and Licandro, Roxane and Ji, Hui and Siddiquee, Md Mahfuzur Rahman and Xu, Daguang and Myronenko, Andriy and Liu, Hao and Pei, Yuchen and Wang, Lisheng and Peng, Ying and Xie, Juanying and Zhang, Huiquan and Dong, Guiming and Fu, Hao and Wang, Guotai and Rieu, ZunHyan and Kim, Donghyeon and Kim, Hyun Gi and Karimi, Davood and Gholipour, Ali and Torres, Helena R and Oliveira, Bruno and Vila{\c{c}}a, Jo{\~a}o L and Lin, Yuzhe and Avisdris, Netanell and Ben-Zvi, Ophir and Bashat, Dafna Ben and Fidon, Lucas and Aertsen, Michael and Vercauteren, Tom and Sobotka, Daniel and Langs, Georg and Aleny{\`a}, Guillem and Villanueva, M{\'o}nica Iglesias and Camara, Oscar and Fadida, Batel S and Joskowicz, Leo and Weibin, Liu and Yi, Liu and Xuesong, Li and Mazher, Moona and Qayyum, Abdul and Puig, Domingo and Kebiri, Hamza and Zhang, Zifan and Xu, Xiaosong and Wu, Di and Liao, Kangle and Wu, Yuxuan and Chen, Jue and Xu, Yongjing and Zhao, Li and Vasung, Lana and Menze, Bjoern and Cuadra, Meritxell Bach and Jakab, Andras}
}

@article{niftymic,
title = {An automated framework for localization, segmentation and super-resolution reconstruction of fetal brain MRI},
journal = {NeuroImage},
volume = {206},
pages = {116324},
year = {2020},
issn = {1053-8119},
author = {Michael Ebner and Guotai Wang and Wenqi Li and Michael Aertsen and Premal A. Patel and Rosalind Aughwane and Andrew Melbourne and Tom Doel and Steven Dymarkowski and Paolo {De Coppi} and Anna L. David and Jan Deprest and Sébastien Ourselin and Tom Vercauteren}
}

@article{antspy,
  title={The ANTsX ecosystem for quantitative biological and medical imaging},
  author={Tustison, Nicholas J and Cook, Philip A and Holbrook, Andrew J and Johnson, Hans J and Muschelli, John and Devenyi, Gabriel A and Duda, Jeffrey T and Das, Sandhitsu R and Cullen, Nicholas C and Gillen, Daniel L and others},
  journal={Scientific Reports},
  volume={11},
  number={1},
  pages={9068},
  year={2021},
}

@article{clinicalGA1,
author = {Papageorghiou, A. T. and Kennedy, S. H. and Salomon, L. J. and Ohuma, E. O. and Cheikh Ismail, L. and Barros, F. C. and Lambert, A. and Carvalho, M. and Jaffer, Y. A. and Bertino, E. and Gravett, M. G. and Altman, D. G. and Purwar, M. and Noble, J. A. and Pang, R. and Victora, C. G. and Bhutta, Z. A. and Villar, J. and { for the International Fetal and Newborn Growth Consortium for the 21st Century (INTERGROWTH-21st) }},
title = {International standards for early fetal size and pregnancy dating based on ultrasound measurement of crown–rump length in the first trimester of pregnancy},
journal = {Ultrasound in Obstetrics \& Gynecology},
volume = {44},
number = {6},
pages = {641-648},
eprint = {https://obgyn.onlinelibrary.wiley.com/doi/pdf/10.1002/uog.13448},
year = {2014}
}

@article{clinicalGAerror,
author = {G. E. Chalouhi and J. P. Bernard and G. Benoist and B. Nasr and Y. Ville and Laurent J. Salomon},
title = {A comparison of first trimester measurements for prediction of delivery date},
journal = {The Journal of Maternal-Fetal \& Neonatal Medicine},
volume = {24},
number = {1},
pages = {51--57},
year = {2011},
publisher = {Taylor \& Francis},
}

\end{document}